\documentclass[10pt,twocolumn,letterpaper]{article}

\usepackage{mpi}
\usepackage{times}
\usepackage{epsfig}
\usepackage{graphicx}
\usepackage{amsmath}
\usepackage{amssymb}
\usepackage{color}
\usepackage{comment}
\usepackage{url}
\usepackage{multirow}
\usepackage{enumitem}

\usepackage[pagebackref=true,breaklinks=true,colorlinks,bookmarks=false]{hyperref}
\newcommand{\norm}[1]{\left\lVert#1\right\rVert}

\begin{document}

\title{Differentiable Event Stream Simulator for Non-Rigid 3D Tracking} 

\author{ 
\begin{tabular}{ccc}
    Jalees Nehvi$^{1,2}$ & Vladislav Golyanik$^{2}$ & Franziska Mueller$^{3}$ \vspace{5pt}\\
    Hans-Peter Seidel$^{2}$ & Mohamed Elgharib$^{2}$ & Christian Theobalt$^{2}$ 
    \vspace{11pt}\\
    $^{1}$Saarland University & $^{2}$MPI for Informatics, SIC & $^{3}$Google Inc.
\end{tabular}
}

\mpifinalcopy 
\maketitle

\begin{abstract}
This paper introduces the first differentiable simulator of event streams, \textit{i.e.,} streams of asynchronous  brightness change signals recorded by event cameras. 
Our differentiable simulator enables non-rigid 3D tracking of deformable objects (such as human hands, isometric surfaces and general watertight meshes) from event streams by leveraging an analysis-by-synthesis principle. 
So far, event-based tracking and reconstruction of non-rigid objects in 3D, like hands and body, has been either tackled using explicit event trajectories or large-scale datasets.  
In contrast, our method does not require any such processing or data, and can be readily applied to incoming event streams. 
We show the effectiveness of our approach for various types of non-rigid objects and compare to existing methods for non-rigid 3D tracking. 
In our experiments, the proposed energy-based formulations  outperform competing RGB-based methods in terms of 3D errors. 
The source code and the new data are publicly  available\footnote{\url{http://gvv.mpi-inf.mpg.de/projects/Event-based_Non-rigid_3D_Tracking}}. 
\end{abstract}


\section{Introduction} 
Template-based non-rigid 3D tracking using a single monocular RGB camera is a well-studied and challenging problem \cite{Perriollat2011, Ngo2015, Yu2015}, due to the ambiguities associated with the monocular setting (\textit{i.e.,} occlusions, lost depth information and many-to-one mapping from the 3D space to the 2D image plane). 
In contrast, 3D tracking non-rigidly deforming shapes from a single monocular event camera is an emerging research field. 
To the best of our knowledge, \textit{there has been no method shown in the literature so far which can track general objects in 3D from a single event stream and without prior knowledge of the observed object class} (in contrast to EventCap  \cite{EventCap2020} which assumes human bodies). 
While conventional cameras record images synchronously, \textit{i.e.,} at regular intervals (\textit{e.g.,} every $33$ ms) and full or partial resolution of the physical pixel matrix, event cameras react to brightness changes asynchronously in space and time. 
Each signal, called \textit{an event}, reports a change of the per-pixel brightness by a certain threshold; it  contains the pixel coordinates of the event occurrence, a timestamp and polarity. 
Thanks to the asynchronous operational principle, timestamp resolution of modern event cameras reaches $1$ $\mu s$. 
Polarity is a binary flag showing whether the brightness at a  given pixel has increased or decreased. 
Advantages of event cameras include high dynamic range, ultra-high  temporal resolution and lower data throughput than the throughput of high-speed RGB cameras recording fast moving and deforming objects. 
Thus, compared to the monocular RGB-based case, 3D tracking  from an event camera poses \textbf{additional challenges}. 
Not only because the colour information is lost but also due to the different data modality, the existing methods for monocular template-based 3D reconstruction cannot be applied to event streams. 
While some principles developed over the decades for the RGB-based setting can be extrapolated to event streams in order to make the problem well-posed (\textit{e.g.,} the assumptions of isometric deformations and volume preservation), entirely new techniques are required to guide the reconstruction. 
Talking about energy-based methods, this means that the data term and geometric regularisers have to be questioned and re-invented for the new and more challenging setting and data modality. 

This paper proposes \textit{a new differentiable event stream simulator for problems on event streams involving non-rigidly deforming objects}. 
The core idea of our technique is to correlate a synthetically  generated event stream pattern with an observed one. 
\textit{We combine computer graphics rendering techniques with the event stream formation model in a differentiable way and obtain a novel analysis-by-synthesis approach.}  
Assuming the initial 3D shape state is known and accurately projects to the image plane---which is the assumption of many monocular template-based non-rigid 3D tracking methods \cite{Yu2015}---we iteratively update the 3D geometry until the induced event stream pattern strongly correlates with the observed event stream pattern. 
One of the advantages of such an approach is that it is correspondence-free, \textit{i.e.,} it does not rely on spatio-temporal event association, unlike \cite{EventCap2020}.  
Second, it works entirely in the event space and does not require complementary greyscale images, unlike \cite{EventCap2020}. 
Third, our component can be readily used in modern neural approaches and, potentially, other problems involving event streams. 
To summarise, the main \textbf{contributions} of this paper are as follows: 
\begin{itemize}%
    \item The first differentiable event stream simulator for problems on event  streams using analysis-by-synthesis and event stream correlation approach. 
    \item The first energy-based method for  template-based  tracking of deformable surfaces in  3D from a single event stream relying on our  differentiable event stream simulator. 
    We show variants of its energy function for  parametric shapes and general meshes. 
    Our method is fully unsupervised and does not  require training data. 
\end{itemize} 
In the experiments with three types of non-rigid objects  (hands, thin surfaces and watertight meshes), we demonstrate  that our method reaches 3D tracking accuracy comparable to  what is achievable by modern RGB-based methods. 
The proposed approach is more accurate at tracking fast deformations, which cause motion blur when recorded 
with a conventional RGB camera. 
Moreover, in several cases, our event-based formulation also outperforms RGB-based methods in the case of  moderately evolving surfaces. 

\section{Related Work} 

\noindent\textbf{Monocular 3D Non-Rigid Tracking.} 
Methods for 3D reconstruction of non-rigid surfaces from  monocular videos can be  classified into non-rigid structure from motion (NRSfM) and template-based approaches. 
NRSfM relies on correspondences across the input views relative to a single keyframe, which are factorised into camera poses and non-rigid shapes for every input frame. 
Whereas earlier methods focused on the  factorisation of sparse tracked keypoints  \cite{Bregler2000, Torresani2008}, recent methods support dense optical flow as input \cite{Fragkiadaki2014, Kumar2018, Golyanik_2019, Sidhu2020}. 
Template-based methods assume that an accurate %
initial 3D state for one sequence frame is known \cite{Gumerov2004, Salzmann2007, Perriollat2011, Oestlund2012, Ngo2015, Yu2015}. 
Several recent methods encode templates in weights of neural networks, which are trained to regress 3D surfaces directly from 2D images \cite{Shimada2019,  Tsoli_2019_ICCV}.

\noindent\textbf{Event-Based 3D Reconstruction.} 
Since event cameras became %
accessible for the broad research community, 
they were applied to various low- and mid-level computer vision tasks \cite{Gallego2020}. 
Several methods have been proposed for 3D reconstruction and tracking of rigid objects in 3D \cite{Schraml2015, Kim2016, Reinbacher2017, Rebecq2018}. 
In contrast, 3D reconstruction of non-rigid objects from a single event stream remains a starkly under-explored problem in the literature. 
Calabrese \textit{et al.}~\cite{Calabrese2019} propose a method for 3D human pose estimation from multiple synchronous event streams. 
EventCap of Xu \textit{et al.}~\cite{EventCap2020} is a hybrid approach that relies on deblurred greyscale images recorded at usual frame rates and an event stream from DAVIS240C to track a rigged 3D model of an actor. 
The recent work of Rudnev \textit{et al.}~\cite{rudnev2020eventhands}
proposes \hbox{EventHands}, 
\textit{i.e.,} a neural method for 3D hand reconstruction from a single event stream. 
In contrast to EventHands, our method does not require large corpora of training data, and works for general non-rigid objects with hands only being one example object. 
All in all, we propose a new way of solving tracking problems from a single event stream in an analysis-by-synthesis fashion, which can be potentially used for other problems than 3D tracking in future. 

Our event correlation approach relates to the method of Bryner \textit{et al.}~\cite{Bryner2019}, who determine an event camera pose with respect to a rigid 3D reconstruction of an environment. 
Their objective function minimises the difference between the measured  (\textit{i.e.,} obtained by integration of real events) and predicted (\textit{i.e.,} obtained by rendering the 3D map) intensity images. 
In contrast, we compare event frames, \textit{i.e.,} observed and predicted, and support deformable objects. 

While there are existing event stream  simulators such as ESIM \cite{Rebecq18corl},  we implement own %
lightweight component that is fully differentiable and easily customisable for target non-rigid 3D tracking and reconstruction applications. 
Note that we do simulate individual events, which are then accumulated in windows for comparisons. 
On the other hand---and as our experiments show---the implemented level of event synthesis fidelity is sufficient for our target applications. 

\section{Event Generation Model and Event Frame} 
\label{sec:thresholding_func}

\noindent\textbf{Event Generation Model.} 
In contrast to RGB cameras which record absolute brightness, event cameras record relative brightness changes in form of asynchronous events.
At a given pixel $\textbf{x}$, an event is triggered if the absolute value of the difference between the incoming brightness received at that pixel and the current brightness stored at that pixel, exceeds a certain threshold called the \textit{contrast sensitivity} of the event camera. 
The pixels are triggered independently, resulting in an asynchronous stream of events, when 
\begin{equation} 
| \mathcal{L}(\textbf{x},t) - \mathcal{L}(\textbf{x}, t-\Delta t) |   \geq C \:,
\label{eq:events}
\end{equation}
where $\mathcal{L}$ represents the brightness, $t$ represents the current timestamp and $C$ represents the event camera threshold.
\noindent\textbf{Event Frame.}
An event is an indicator of brightness change at a given pixel at a particular instant of time. 
It is usually represented as a tuple $e = (x,y,t,p)$, where $x$ and $y$ denote the pixel coordinates, $t$ the timestamp and $p \in \{-1,1\}$ the polarity which indicates whether there has been a positive or negative change in brightness. 
Since individual events carry little information, we accumulate several events over a span of time.
Let $\mathcal{S} = \{e_i = (x_i, y_i, t_i, p_i)\}$ be the set of events that occurred within a time window $T$.
We consolidate these events into a single \textit{event frame}
\begin{equation}
\mathcal{E}(\textbf{x}) = \sum_{e_i \in \mathcal{S}} p_i \: \delta((x_i,y_i)-\textbf{x}) \:, \label{eq:input_events}
\end{equation}
where $\delta$ is the impulse function.
$T$ is set such that there is enough information per event frame and the high temporal resolution property of the event camera is not compromised.
The event frames are normalised in the range  $[-1,1]$. 

\section{Overview} 
\label{sec:Overview} 
Our goal is 3D tracking of non-rigid objects from event streams.
Since our differentiable event simulator is independent of the exact parameterisation of the object of interest, it can be used with parametric 3D morphable models as well as general 3D meshes. 
For each event frame, we aim to estimate the current 3D geometry and global rigid transform, consisting of translation $t$ and rotation $R$, such that the generated event frame is aligned with the incoming event frame.
For a parametric model (\textit{e.g.}, hands  \cite{Romero2017, Qian2020}), the 3D geometry is controlled by the pose parameters $\theta$, yielding the set of parameters $\Theta = (\theta,t,R)$. 
For general 3D meshes, we assume the availability of a template mesh and express the 3D geometry as displacement field of the vertices. In this scenario, the set of parameters we optimise is $\Theta = (\textbf{V},t,R)$, where $\textbf{V} = \{v_{i}\}_{i=1}^N$ are the 3D coordinates of mesh vertices and $N$ is the total number of vertices.

\begin{figure*}[t]
    \centering{ \includegraphics[width =0.95\textwidth,trim={0 6cm 2.5cm 0},clip]{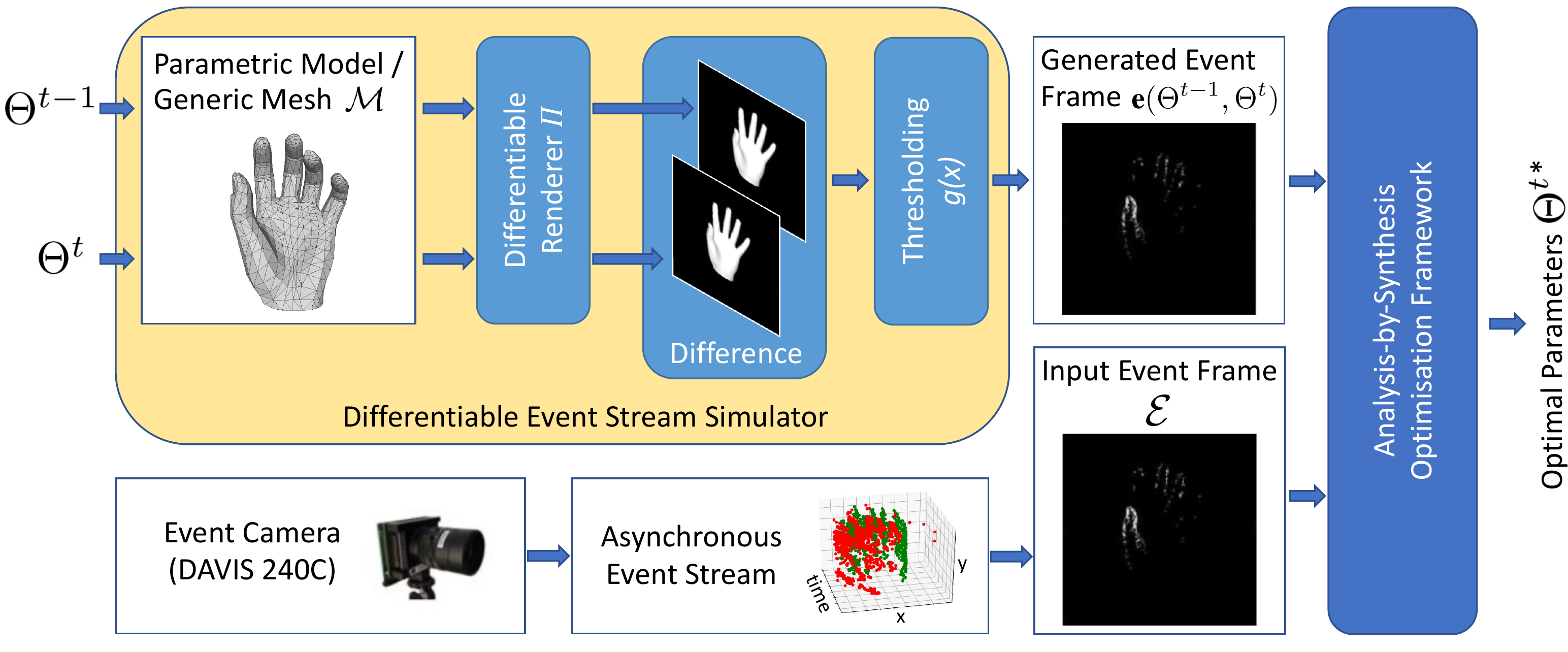}}
    \caption[Method Pipeline]{\textbf{Our non-rigid 3D tracking framework.} We propose a new differentiable event simulator to generate event frames for the current pose hypothesis $\Theta$. We then optimise the parameters $\Theta$ by comparing the generated event frame to the input event frame in an analysis-by-synthesis optimisation framework. The positive and negative events of the input stream are shown in green and red, respectively.
    } 
    \label{fig:pipeline} 
\end{figure*}
Given the input event frames and the parametric model or a mesh, we start from an initial parameter set $\Theta$ 
and render an image of the mesh. 
Our goal is to find a parameter set such that the events generated by an \textit{event generation model} (see Sec.~\ref{sec:thresholding_func}) match the incoming events. 
We optimise the unknown parameters by minimising the objective function in a frame-wise manner, see \eqref{eq:3} and \eqref{eq:objective_generic_mesh}.  
After we have found the optimised parameters for the first event frame, 
we use them as an initialisation for the next event frame and proceed in a similar manner for subsequent frames. 

In our experiments, we use the DAVIS 240C event camera which captures both intensity images as well as event streams of $240{\times}180$ pixels with temporal resolution in the microsecond range.
The intrinsic camera parameters are known. 
Note that our method is purely based on the asynchronous event stream input, and hence we do not use the complementary greyscale images of DAVIS 240C. 

\section{Method}\label{sec:method} 
An overview of our framework is given in Fig.~\ref{fig:pipeline}.
We want to find the optimal parameters $\Theta^*$ by comparing the event frame generated with the current parameter hypothesis $\Theta$ to the input event frame.
First, we use a differentiable renderer to render two images using the meshes from the current and previous timestamps.
We then calculate the per-pixel differences and threshold them according to the event generation model. 
To ensure the differentiability of the thresholding process, we use a differentiable approximation of the staircase function.
Finally, the generated events are used in objective function terms which compare the incoming events with the generated events (\textit{i.e.}, by means of signal correlation).

\subsection{Differentiable Event Simulator}

The main component of our framework is the differentiable event simulator which generates the event frame corresponding to the current parameter hypothesis $\Theta$.
This synthesised event frame is later used for optimising an analysis-by-synthesis objective function.
To obtain the event frame, we render two greyscale images, one corresponding to the mesh at the previous time stamp and one corresponding to the mesh at the current parameter hypothesis $\Theta$.
We make use of differentiable rendering such that the images are differentiable 
with respect to the parameters $\Theta$ controlling the mesh deformations.
We normalise the rendered images in the range $[0, 1]$, which is based on an empirical observation and leads to stable gradients. 
We assume that the object's surface is Lambertian and the lighting conditions are constant.
The rendering is followed by the subtraction of the previously rendered image from the current one. 
Finally, the difference image is thresholded using the smooth thresholding function discussed below, resulting in a generated event frame.
\noindent\textbf{Thresholding Function.} We %
approximate the inequality operation in the event generation model in the form of a smooth threshold function which is a trade-off between the hyperbolic tangent function and the staircase function: 
\begin{equation}
g(x) = \bigg(\frac{x+\epsilon}{|x|+\epsilon}\bigg)\bigg(\frac{1}{1+e^{-w|x|+wC}}\bigg), 
\label{eq:sim}
\end{equation}
where $x$ represents the difference image, $w$ denotes the weight that controls the smoothness of the curve, $C$ represents the threshold or contrast sensitivity of the event camera and $\epsilon$ is the tolerance which ensures stable gradients. 
In general, all the aforementioned steps involved in event frame generation can be mathematically formulated as 
\begin{equation}
\textbf{e}(\Theta^{t-1},\Theta^t) = g(\Pi(\mathcal{M}(\Theta^t)) - \Pi(\mathcal{M}(\Theta^{t-1}))), %
\label{eq:gen}
\end{equation} 
where $\textbf{e}$ is the generated event frame;  $\Theta^t$ and $\Theta^{t-1}$ represent the current parameter set and the parameters estimated for the last frame, respectively; $\mathcal{M}$ is the parametric model or generic mesh; $\Pi$ denotes the image rendering operation and \textit{g} is the smooth threshold function defined in \eqref{eq:sim}. 
The generated and input event frames 
are then fed into an optimisation framework  minimising objective functions---which we describe in Secs.~\ref{sec:handsobjective} and \ref{sec:meshesobjective}---to estimate the unknown optimal parameters $\Theta^*$ in a frame-wise manner. 

\subsection{Objective Function: Parametric Model}
\label{sec:handsobjective}
In the case of a parametric model like hands 
(we use MANO \cite{Romero2017}), the objective comprises of two main terms, \textit{i.e.,} a data term and a regularisation term: 
\begin{equation}
 E(\Theta) = E_{\textit{data}}(\Theta) + \lambda E_{\textit{reg}}(\Theta) \:,
 \label{eq:3}
\end{equation}
where $\lambda$ is a regularisation hyperparameter and $\Theta = (\theta,t,R)$ (see Sec.~\ref{sec:Overview}) represents the set of all parameters, consisting of model parameters for pose $\theta$ and the rigid transform defined by translation $t$ and rotation $R$. 

\subsubsection{Data Term}

To account both for the occurred and missing changes in  the observed 3D shape, we split the data term into an  \textit{event} and a \textit{no-event} terms: 
\begin{equation}
 E_{\textit{data}}(\Theta) =  E_{\textit{event}}(\Theta) +  \lambda_{1}E_{\textit{no-event}}(\Theta) \:,
 \end{equation}
where $\lambda_{1}$ %
balances the two data terms. 

\noindent\textbf{Event Term:} 
The event term penalises the difference between the input event frame $\mathcal{E}$ and the generated event frame (Eq.~\ref{eq:gen}), at only those pixel locations where an event is present in the input event frame.
\begin{equation}E_{\textit{event}}(\Theta) = \sum_{\textbf{x} \in \Omega}(\gamma(\textbf{x}) \big( \mathcal{E}(\textbf{x}) -    \textbf{e}(\Theta^{t-1},\Theta)(\textbf{x}))\big)^{2} \:,
\label{eq:event_term}
\end{equation}
where $\Omega$ denotes the spatial image domain and $\gamma(\cdot)$ is an indicator function that yields $1$ if an event occurred at pixel $\textbf{x}$ in the input event frame and $0$ otherwise. 

\noindent\textbf{No-Event Term:}
On the contrary, the no-event term penalises any non-zero value present in the generated event frame at the corresponding locations in the input event frame where an event is absent: 
\begin{equation}E_{\textit{no-event}}(\Theta) = \sum_{\textbf{x} \in \Omega}\big(\bar{\gamma}(\textbf{x})(\textbf{e}(\Theta^{t-1},\Theta)(\textbf{x}))\big)^{2} \, ,
\end{equation}
where $\bar{\gamma}(\textbf{x}) = 1 - \gamma(\textbf{x})$ indicates pixels where no event occurred in the input event frame.
Note that without $E_{\textit{no-event}}$, the tracked 3D  shape would not be sufficiently constrained. 
\subsubsection{Regularisation Term}
The regularisation term ensures temporal smoothness by penalising large parameter changes between the current and the previous event frame.
This helps to reduce flickering of the estimated mesh between adjacent frames: 
\begin{equation}
E_{\textit{reg}}(\Theta) = \|\Theta - \Theta^{t-1}\|_{2}^{2},  \label{eq:regularizer_parametric}
\end{equation}
where $\Theta$ is the current parameter set and $\Theta^{t-1}$ is the parameter set estimated for the previous frame.
\subsection{Objective Function: Meshes}
\label{sec:meshesobjective}
Instead of a parametric model, here we are provided with a generic mesh parameterised by $\Theta = (\textbf{V}, t, R)$ where $\textbf{V}$ is the set of mesh vertices, $t$ is global translation, and $R$ is global rotation (see Sec.~\ref{sec:Overview}). 
The objective function comprises of the following terms: 
\begin{multline}E(\Theta) = E_{\textit{data}}(\Theta)+ 
\lambda_{\textit{top}} E_{\textit{top}}(\Theta)+\lambda_{\textit{iso}} E_{\textit{iso}}(\Theta)\\
+\lambda_{\textit{geo}} E_{\textit{geo}}(\Theta) + \lambda_{\textit{reg}} E_{\textit{reg}}(\Theta), 
\label{eq:objective_generic_mesh}
\end{multline}
where $\lambda_{\textit{top}},\lambda_{\textit{iso}},\lambda_{\textit{geo}}$ and $\lambda_{\textit{reg}}$ are 
scalar weights. 
The data term is further split into an event term and a silhouette term: 
\begin{equation}
 E_{\textit{data}}(\Theta) =  E_{\textit{event}}(\Theta) +  \lambda_{\textit{sil}}E_{\textit{sil}}(\Theta), 
 \end{equation}
where $\lambda_{\textit{sil}}$ is another hyperparameter that balances the importance of each term.
The event and regularisation terms remain the same as in \eqref{eq:event_term} and \eqref{eq:regularizer_parametric}, respectively. 
We now discuss the new terms of the objective function in detail. 

\subsubsection{Silhouette Term}
The silhouette term matches the relevant events with the 2D projection of the nearest visible mesh vertices. 
The nearest vertices are calculated based on the shortest Euclidean distance between an event and the vertices in the 2D image:
\begin{equation}
E_{\textit{sil}}(\Theta) = \sum_{e_b \in \hat{\mathcal{E}}}||\pi(v_{b}) - (x_b, y_b)||_{2}^{2}, 
\end{equation}
where $v_{b}$ is the 3D mesh vertex closest to the event $e_b$ in the 2D space, $(x_b, y_b)$ represents the 2D location of $e_b$, $\pi$ denotes the 3D-to-2D projection and $\hat{\mathcal{E}}$ stands for the set of relevant events in the current event frame. 
The noisy events are filtered out by taking a $5{\times}5$ neighbourhood around each event in the event frame and then removing them if the number of events present in this region is below some threshold. 
\subsubsection{Topology-Preserving Term}
The topology-preserving term demands that the relative positions of the current mesh vertices with respect to their neighbours remain the same as those of the template mesh, \textit{i.e.,} it ensures spatial  smoothness of the surface: 
\begin{equation}
E_{\textit{top}}(\Theta) = \sum_{\textbf{i} = 1}^{N} \sum_{\textbf{j} \in \mathcal{N}_{i}} || (v_{i} - v_{j}) - (\textbf{v}_{i} - \textbf{v}_{j})||_{2}^{2}, 
\end{equation}
where $N$ represents the total number of mesh vertices,  $v_{i}$ denotes the $i$-th vertex of the current mesh, $\textbf{v}_{i}$ is the $i$-th vertex of the template mesh and $j \in \mathcal{N}_{i}$ indexes the points in the  neighbourhood of the $i$-th vertex. 
\subsubsection{Isometric and Geodesic Terms}%
We consider isometric shape deformations, \textit{i.e.}, that the area of the mesh or the distance between any two points on the surface of the mesh remain unchanged.
Thus, the isometric term ensures that the mesh edge length between neighbouring vertices is preserved with respect to the template.
The geodesic term, similarly, ensures that the distance between any two non-neighbouring vertices on the mesh surface remains preserved with respect to the template.
Since computing this distance for all possible combinations of vertices can be computationally very expensive, we, therefore, uniformly sample vertex points on the template mesh surface in a coarse manner (we take every tenth point) and compute the geodesic loss for this grid of sparse points only: 
\begin{equation}
E_{\textit{iso}}(\Theta) = \sum_{\textbf{i} = 1}^{N} \sum_{\textbf{j} \in \mathcal{N}_{i}} \big(||v_{i} - v_{j}|| - ||\textbf{v}_{i} - \textbf{v}_{j}||\big)^{2}, 
\end{equation}
\begin{equation}
E_{\textit{geo}}(\Theta) = \sum_{\textbf{i} \in \Omega} \sum_{\substack{\textbf{j} \in \Omega,\\j\neq i}} \big(d(v_{i},v_{j}) - d(\textbf{v}_{i},\textbf{v}_{j})\big)^{2}, 
\end{equation}
where $\Omega$ denotes the grid of sub-sampled vertices and $d$ denotes the geodesic distance between two points. 
\subsection{Initialisation}
Depending on whether we have a generic mesh or a morphable model, we either initialise with the template mesh or the mean shape of the model.
For the initial frame, we start from a known position.
Regarding subsequent event frames, we initialise with the estimated parameters of the previous frame and optimise for all the parameters.
For each frame, we first perform rigid fitting, \textit{i.e.}, optimising for $t$ and $R$ only keeping the rest of the parameters fixed, and thereafter we optimise for the other parameters.
\subsection{Implementation Details}
We implement our method in \textit{PyTorch}  \cite{pytorch2019}.  
The optimiser used is \textit{Adam} \cite{Kingma2014} with step size of $5 \cdot 10^{-4}$. 
The method 
takes ${\approx}1.5$ minutes per frame on NVIDIA V100 GPU. 
We balance the various energy terms with $\lambda = 10,\ \lambda_{1}= 0.1,\ \lambda_{sil}=0.01,\ \lambda_{top} = 1,\ \lambda_{iso}=1 ,\ \lambda_{geo} =1,$ and $\ \lambda_{reg} = 10$.

\section{Experimental Results}

\begin{figure*}[h]
    \centering{\hspace*{0.085cm} \includegraphics[width =1.0\textwidth]{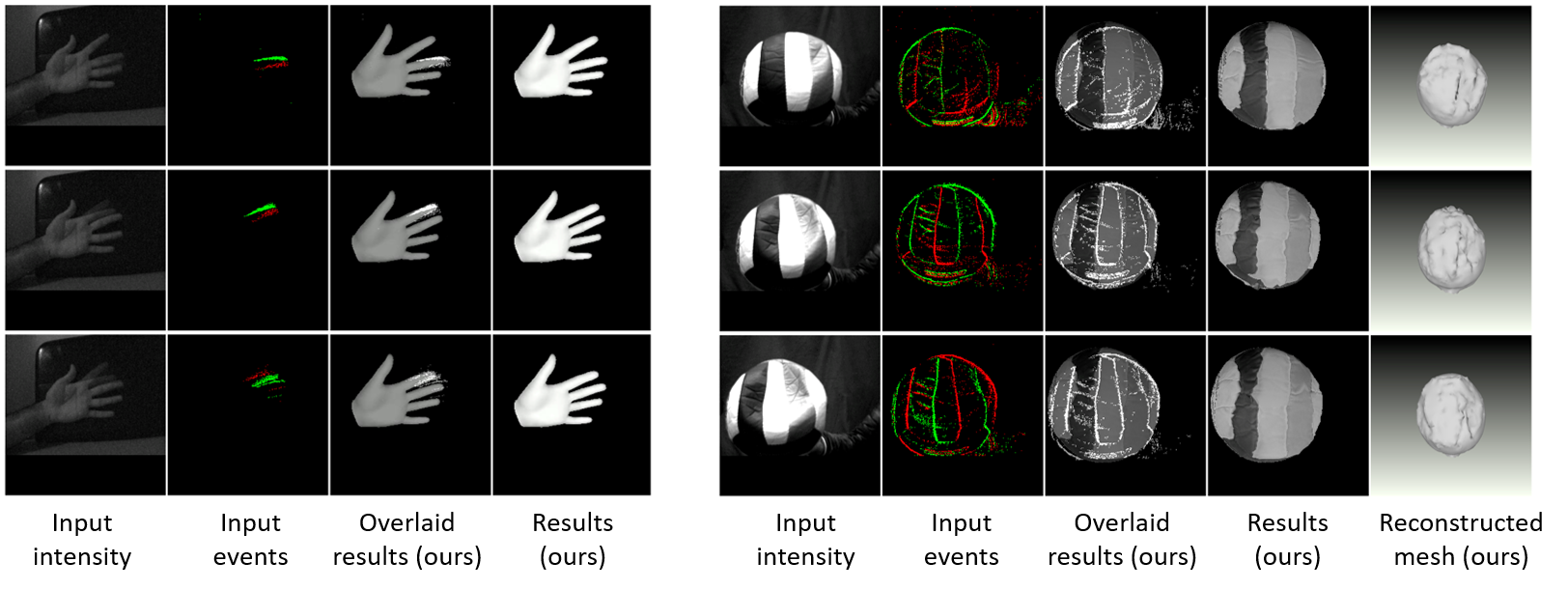}}
    \caption[Qualitative Results]{Results produced by our technique on the real hand (left) and real ball (right) sequences.} 
    \label{fig:qualitativerealhandball} 
\end{figure*}
\begin{figure*}[h]
    \centering{\hspace*{-0.05cm} \includegraphics[width =1.0\textwidth]{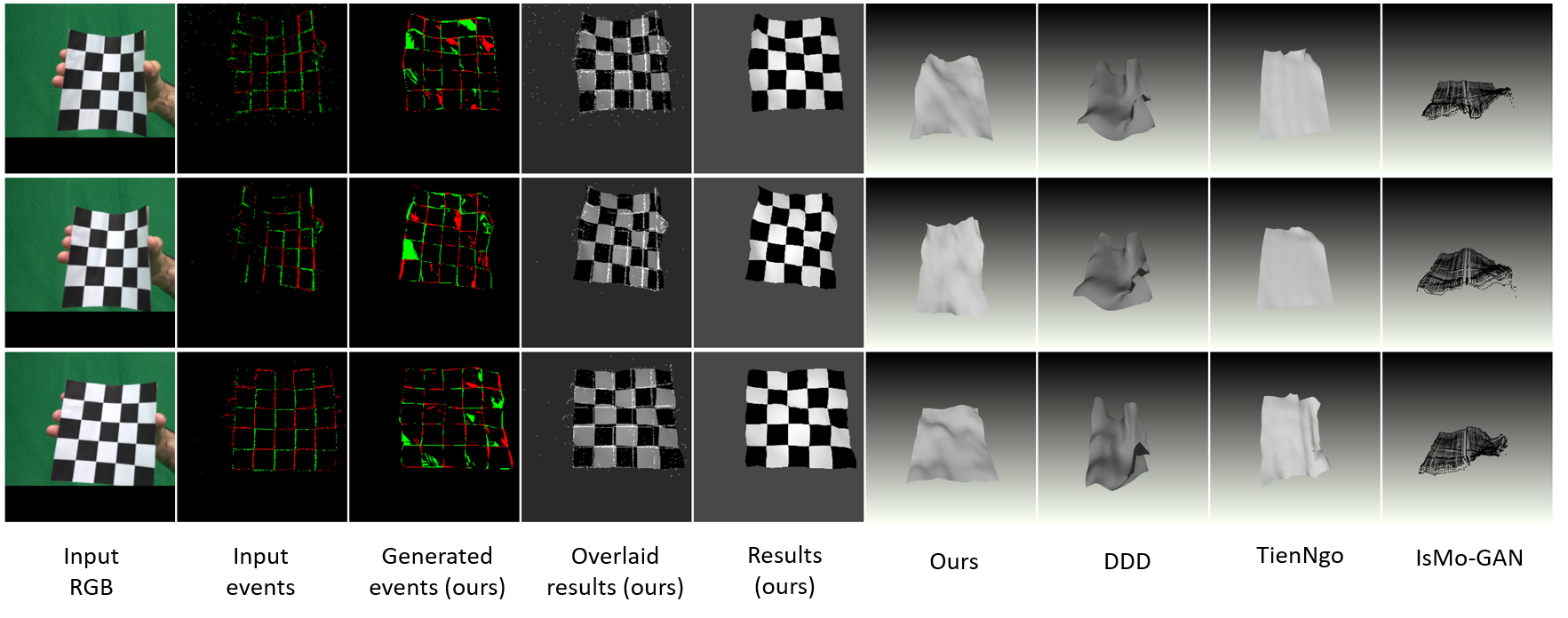}}
    \caption[Qualitative Results]{Results on the \textit{real paper}. 
    Ours produces more feasible reconstructions and fewer surface artefacts in the comparisons. %
    } 
    \vspace{-5pt} 
    \label{fig:qualitativerealpaper} 
\end{figure*}
In this section, we report our experiments with the proposed non-rigid tracking methods using our differentiable event stream simulator. 
We consider three types of non-rigid objects in our  experiments: a hand mesh, a deformable sheet (paper) and a spherical watertight mesh (ball). 
For hands, we use a 3D morphable hand model \cite{Romero2017}. 
We compare to several existing state-of-the-art methods for RGB-based non-rigid tracking, including template-based methods of Tien Ngo \textit{et al.}~\cite{Ngo2015} and  Direct-Dense-Deformable (DDD) \cite{Yu2015}, learning-based method Isometric Monocular GAN (IsMo-GAN) \cite{Shimada2019} and HandGraphCNN of Ge \textit{et al.} \cite{ge2019handshapepose} for 3D hand shape reconstruction. 
To ensure accurate initialisation for template-based methods (\textit{i.e.,} in the first frames), we perform manual adjustment, if required. 

\subsection{Evaluation Metrics} 
For hands, we use the average 3D joint error as our metric for  quantitative evaluation: 
\begin{equation}
e_{joint3D} = \frac{1}{N}\sum_{i=1}^{N}\frac{\norm{J_{GT}^{i} -  J_{rec}^{i}}_F}{\norm{J_{GT}^{i}}_F}, 
\end{equation}
where $N$ is the number of frames and $\norm{\cdot}_F$ denotes Frobenius norm; $J_{GT}$ and $J_{rec}$ denote the ground-truth and reconstructed 3D joint locations, respectively. 
For meshes, we employ average $e_{3D}$ which differs from $e_{joint3D}$ in that it compares the dense reconstructed meshes $S_{rec}$ with ground-truth meshes $S_{GT}$ instead of sparse sets of joints. 
Both $e_{joint3D}$ and $e_{3D}$ are reported after Procrustes alignment of the shapes, \textit{i.e.,} with the resolved translational and rotational components. 

\vspace{7pt} 

\subsection{Real Sequences} 
We record real sequences of the three classes of 3D objects: hands, paper (deformable sheet) and ball.
Our setup comprises of DAVIS 240C for recording events and an additional RGB sensor (Sony RX0) that records coloured intensity images at $50$ fps.
The coloured intensity images are used for the methods being compared against our approach, since they are RGB image-based methods.
We calibrate the cameras together and compute the extrinsics. For synchronisation, we use a flash.
For a $36$-seconds-long recording of a scene with significant motion, the events take up around $23$ MB of storage space as compared to the corresponding video which takes up around $110$ MB of space.
We record a $20$-seconds-long video for fast hand motion comprising of around $1500$ event frames; another short $10$-seconds video demonstrating the deformation of a volleyball, comprising of around $400$ event frames and another $50$-seconds-long video of a deforming sheet of paper with printed texture, comprising of $1100$ event frames. 
The spatial resolution of event as well as intensity frames is set to $240{\times}240$ pixels. 
We use $T = 800$ events for hand and $T = 2000$ events for paper and ball, see \eqref{eq:input_events}, and set $C = e^{0.5}$ for real data on the $[0, 255]$ scale. 
Figs.~\ref{fig:qualitativerealhandball}--\ref{fig:qualitativerealpaper} show results of various methods on the real sequences.  
Our technique produces realistic results that overlap well with the input event stream, whereas results of competing methods evince various surface artefacts and folding which are not observed in the input images. 

\begin{figure}[t]
    \centering{\hspace*{-0.55cm} \includegraphics[width =0.5\textwidth]{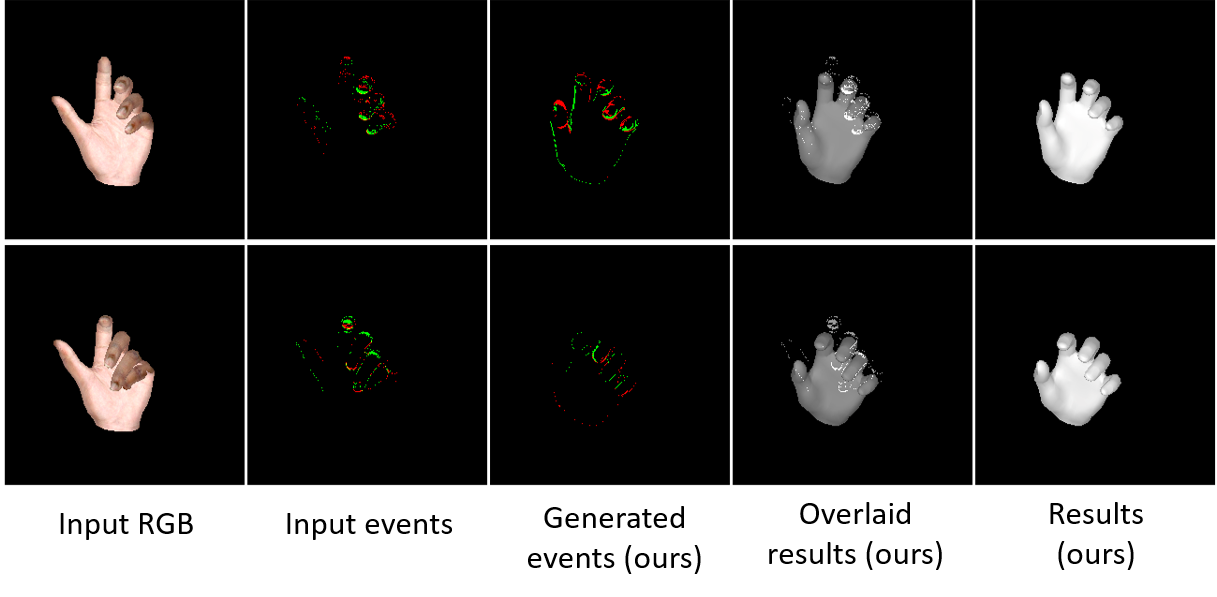}}
    \caption[Qualitative Results]{Results of our technique on the \textit{synthetic hand} sequence.} 
    \vspace{-7pt}
    \label{fig:qualitativesynhand} 
\end{figure}

\begin{figure*}[h]
    \centering{\hspace*{0.095cm}\includegraphics[width =1.0\textwidth]{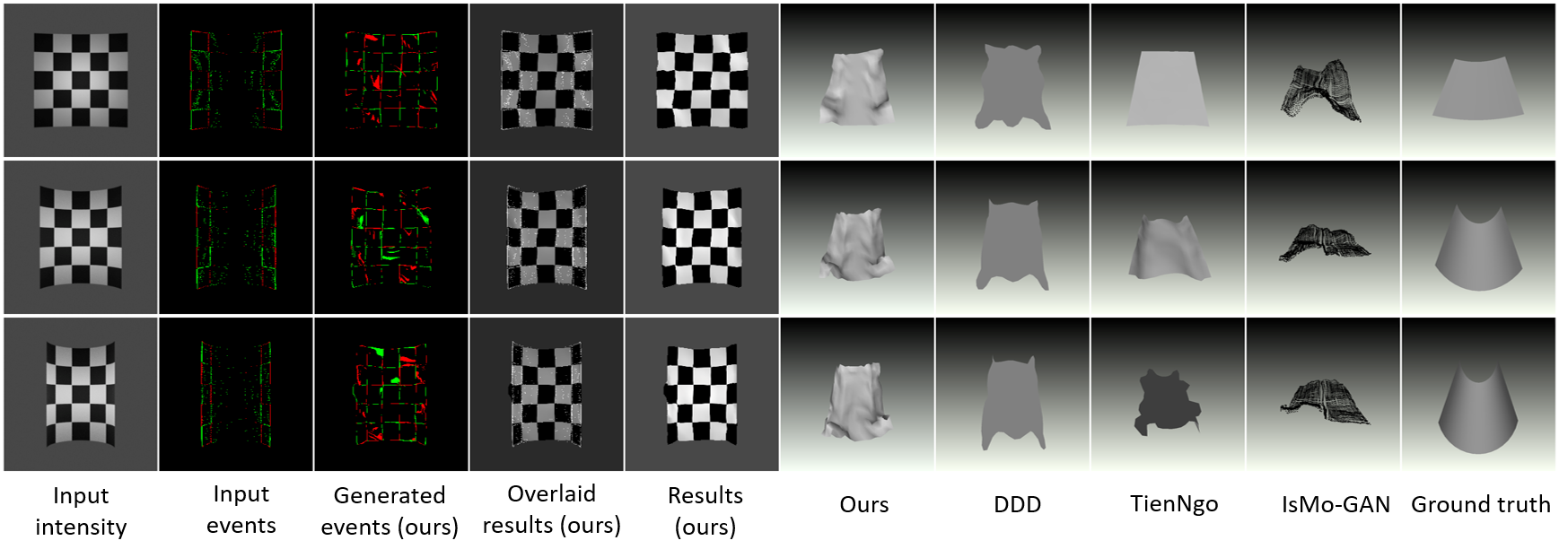}}
    \caption[Qualitative Results]{Results on the \textit{synthetic paper} sequence. Our technique outperforms existing methods quantitatively (see Table~\ref{Tab:t3}).} 
    \label{fig:qualitativesynpaper} 
\end{figure*}
\begin{figure*}[h]
    \centering{ \includegraphics[width =1.0\textwidth]{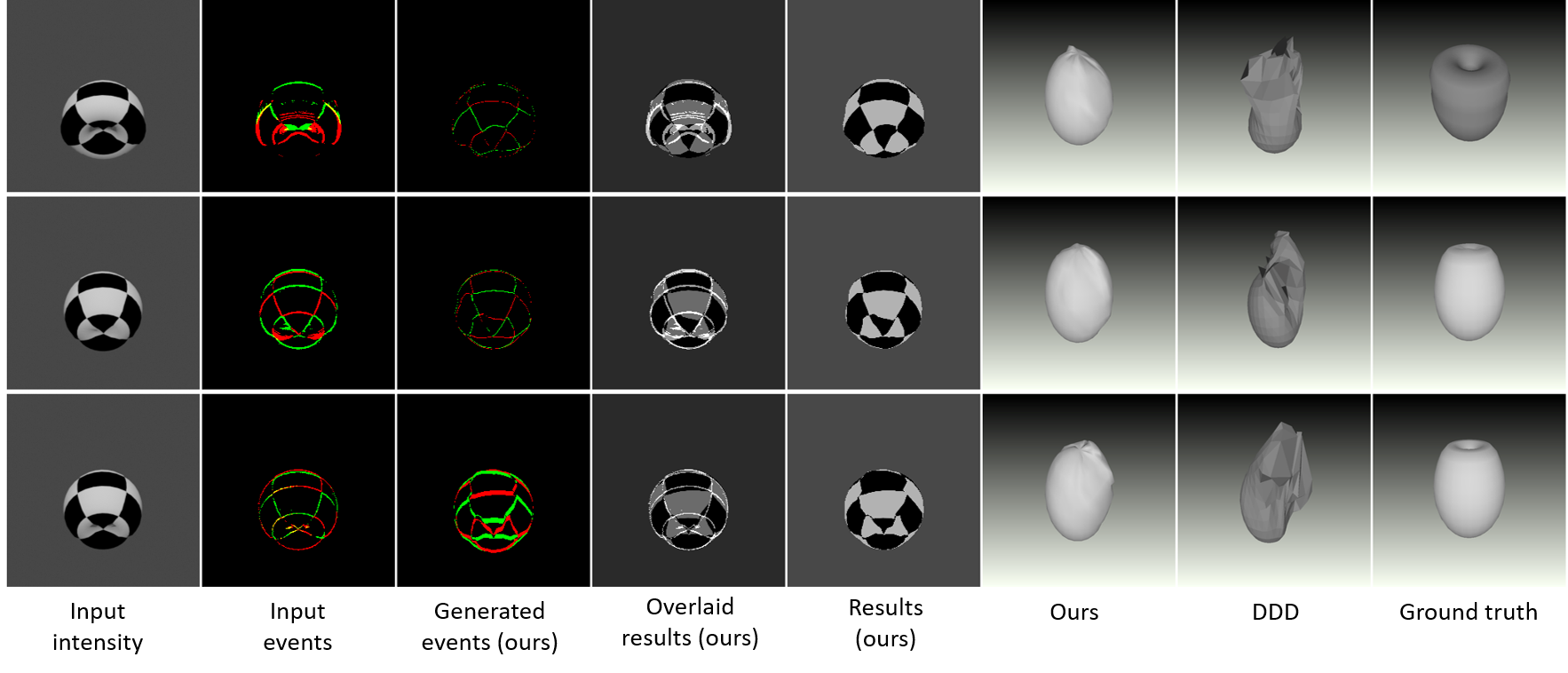}}
    \caption[Qualitative Results]{Results on the \textit{synthetic ball} sequence. Our technique outperforms existing methods quantitatively (see Table~\ref{Tab:t4}).} 
    \label{fig:qualitativesynball} 
\end{figure*}

\subsection{Controlled Experiments} 
\label{sec:numericalanalysis}
We prepare synthetic data with 3D ground truth for quantitative analysis. 
We render synthetic RGB image sequences and event stream data for each object.
In the case of hands, we sample the pose parameter space using adaptive sampling similarly to Rebecq \textit{et  al.}~\cite{Rebecq18corl}, see our supplement for the details and render the corresponding intensity images. 
For the generic meshes, we deform the meshes in Blender  \cite{Blender} and render the intensity images at $50$ fps. 
Then, we generate synthetic event data from two consecutive images by subtracting them, followed by thresholding the difference image. 
The sequences contain $306$ (\textit{hand}), $250$ (\textit{paper}) and $200$ (\textit{ball}) frames. 
We use the following values for the time window $T$ in \eqref{eq:input_events}: 
$400$ events for \textit{hand} and $1200$ events for \textit{paper} and \textit{ball}, and set $C = 10$ for synthetic data. 

For the \textit{synthetic hand} sequence, we report 
$e_{joint3D}$ in Table~\ref{Tab:t2}. 
Our method achieves approximately twice as lower error as HandGraphCNN~\cite{ge2019handshapepose}. 
The latter is a learning-based approach operating on RGB images, on each frame individually, that does not assume accurate shape initialisation for one of the frames. 
Considering that the images, on which HandGraphCNN has been trained, are  different from our rendered images which leads to domain gap, it is  conceivable that HandGraphCNN does not outperform our method. 
Note that we cannot retrain HandGraphCNN on our data, because we render only a few hundreds of images (our method is unsupervised). 
Nevertheless, the strength of our approach comes from its use of the data term which is capable of tracking such highly articulated objects as human hands. 
The visual performance of our technique on the \textit{synthetic hand} sequence is shown in Fig.~\ref{fig:qualitativesynhand}. 

For the \textit{synthetic paper} sequence, the 3D  reconstruction error is reported in  Table~\ref{Tab:t3}. 
Our method is ranked first and we slightly outperform the  method of Tien Ngo \textit{et al.}~\cite{Ngo2015} in $e_{3D}$. %
Both DDD and IsMo-GAN are not able to track the observed  surfaces accurately, though DDD shows a slightly worse  accuracy compared to \cite{Ngo2015}. 
Note that both Tien Ngo \textit{et al.} \cite{Ngo2015} and DDD \cite{Yu2015} do not suffer from domain gap of learning-based methods. 
On the other hand, IsMo-GAN is positioned as a generalisable approach, due to the powerful silhouette term, which allows it to reconstruct surfaces that are significantly different from  those observed in the training set. 
Fig.~\ref{fig:qualitativesynpaper} shows visualisations of the results obtained by the tested methods on the \textit{synthetic paper} sequence. 

For the \textit{synthetic ball} sequence, we report 
$e_{3D}$ in Table~\ref{Tab:t4}. 
We compare to DDD~\cite{Yu2015} which supports watertight  meshes. 
On average, we outperform DDD by ${\approx}30\%$ and our  reconstructions reflect the deformations in the occluded  parts more faithfully. 
The qualitative improvement in the results of our approach over DDD is shown in Fig.~\ref{fig:qualitativesynball}. 
\begin{table}[h]
\small
\begin{center}
\begin{tabular}{|l|c|c|}
\hline
Method & $e_{joint3D}$ & std.~deviation \\
\hline\hline
HandGraphCNN \cite{ge2019handshapepose} & 0.191 & $\pm $0.055\\
Ours & \textbf{0.074} & $\pm $0.027\\ 
\hline
\end{tabular}
\end{center}
\caption{Quantitative results on the \textit{synthetic hand} sequence.}
\label{Tab:t2}
\end{table}
\vspace{-5pt} 
\begin{table}[h]
\small
\begin{center}
\begin{tabular}{|l|c|c|}
\hline
Method & $e_{3D}$ & std.~deviation \\
\hline\hline
DDD \cite{Yu2015}  & 0.266 & $\pm $0.12  \\
Tien Ngo \textit{et al.} \cite{Ngo2015} & 0.235 & $\pm $0.158\\
IsMo-GAN \cite{Shimada2019} & 0.384 & $\pm $0.092\\
Ours & \textbf{0.232} & $\pm $0.135 \\
\hline
\end{tabular}
\end{center}
\caption{Quantitative results on the \textit{synthetic paper} sequence.}
\label{Tab:t3}
\end{table}
\vspace{-5pt}
\begin{table}[h]
\small
\begin{center}
\begin{tabular}{|l|c|c|}
\hline
Method & $e_{3D}$ & std.~deviation \\
\hline\hline
DDD \cite{Yu2015} & 0.656 & $\pm $0.151\\
Ours & \textbf{0.47} & $\pm $0.31\\
\hline
\end{tabular}
\end{center}
\caption{Quantitative results on the \textit{synthetic ball} sequence.}
\label{Tab:t4}
\end{table}

\subsection{Ablation Study}

We perform an ablation study for the proposed energy-based method for non-rigid 3D tracking. 
The summary of the attained $e_{3D}$ is provided in Table  ~\ref{Tab:t1}. 
Here, we evaluate on the same synthetic sequences used in Sec.~\ref{sec:numericalanalysis}.
We observe that using all terms produces the best results. 
Leaving out any of the energy terms leads to an accuracy drop. 
For the \textit{ball} sequence, leaving out the silhouette term or both the isometric and topology preserving terms leads to the most significant accuracy drop. 
For the \textit{hand} sequence, leaving out the no-event term leads to a significant accuracy drop. 
Please note that processing each of these sequences requires different energy terms (see Sec.~\ref{sec:handsobjective} and Sec.~\ref{sec:meshesobjective}). 
\begin{table}[h]
\small
\begin{center}
\begin{tabular}{|c|l|c|c|}
\hline
& Method & $e_{3D}$ & std.~deviation \\ 
\hline\hline
\multirow{5}{*}{\textit{Ball}}& 
All terms & \textbf{0.467} & $\pm $0.312\\
& w/o Silhouette term & 0.571 & $\pm $0.326\\
& w/o Topological term & 0.564 & $\pm $0.307\\
& w/o Isometric terms & 0.546  & $\pm $0.303\\
& w/o Top + Iso terms & 0.579 & $\pm $0.293\\\hline 
\multirow{2}{*}{\textit{Hand}}&
All terms & \textbf{0.074} & $\pm $0.027 \\
& w/o No-Event term & 0.141  & $\pm $0.055 \\
\hline
\end{tabular}
\end{center}
\caption{Ablative analysis on the  \textit{synthetic ball} and \textit{hand}.} 
\label{Tab:t1}
\end{table}

\section{Conclusion}
We introduced the first differentiable event stream simulator which enables non-rigid 3D tracking of deformable objects. 
In contrast to previous event-based tracking techniques, our tracking approach does not require large training datasets or computation of explicit event trajectories. 
Instead, we leverage the differentiability of our event simulator in an analysis-by-synthesis optimisation framework.
We demonstrated the generality of our method by performing quantitative and qualitative evaluation of our method for different deformable objects. 
Future work can address improving the processing speed and making it real-time. 

One important conclusion which we draw from the experiments is that events not only pose additional challenges and add complexity for non-rigid 3D tracking, but they also can improve the tracking accuracy compared to the RGB-based methods. 
This is because events represent a more abstract data modality compared to RGB images, which we leverage with appropriate novel data terms and regularisers. 
We believe the proposed framework can be used for other tasks in event-based vision such as 2D tracking and gesture recognition. 
It can also be used as a differentiable component in supervised learning methods. 
We hope our framework inspires more future work in  event-based vision. 

\noindent\textbf{Acknowledgement.} This work was  supported by the ERC Consolidator Grant 770784. 

{\small
\bibliographystyle{ieee_fullname}
\bibliography{bibliography}
}

\onecolumn 
\newpage 
\twocolumn
\begin{center}
\textbf{{\Large Supplementary Material}} 
\end{center}
\appendix

In this supplement, we provide further details about the proposed event stream simulator 
including adaptive sampling for parametric models   (Sec.~\ref{sec:adaptive_sampling}) and application of our approach in tracking fast motions  (Sec.~\ref{sec:tracking_fast_motions}). 
\textit{See our supplementary video for dynamic visualisations.} 

\section{Adaptive Sampling}\label{sec:adaptive_sampling} 
In the case of parametric models like hands, synthetic event generation involves sampling the pose space parameters followed by rendering the corresponding 3D states. 
The current rendered image and the previously rendered image are subtracted, followed by thresholding. 
Uniformly sampling the parameters has two main drawbacks. 
First, it results in redundancy if the sampling rate is too high since  there is a small change between the two subsequent images. 
Secondly, if the sampling rate is too low, a significant  portion of  relevant event information will be lost. 
We apply adaptive policy \cite{Rebecq18corl} which tackles this problem by sampling the parameter space intervals according to the time stamps as follows: 
\begin{equation}
t_{k+1} = t_{k} + \lambda \, C {\left \lvert max_{\textbf{x} \in \Omega} \frac{\partial \mathcal{L}(\textbf{x};t_{k})}{\partial t} \right \rvert ^{-1}} \: ,
\end{equation}
where $\mathcal{L}$ is the difference image intensity, $t_{k}$ is the current time stamp, $\textbf{x}$ is the pixel location, $t_{k+1}$ is the next time stamp of the adaptive sampling, $\Omega$ is the image plane, $\lambda = 2$ 
is a constant and $C$ is the contrast sensitivity or threshold and is set to $10$.
The derivative $\frac{\partial \mathcal{L}}{\partial t}$ is approximated by dividing the difference image by the length of the previous time interval. 
The new timestamp ensures that very few events are  missed out between two consecutively rendered image  frames and, at the same time, that any two consecutively rendered frames are non-identical. 
\section{Application: Tracking Fast  Motions}\label{sec:tracking_fast_motions} 
Conventional RGB cameras suffer from motion blur while recording fast motions. 
Moreover, the use of high-speed cameras for fast motion capture results in a lot of data which need to be stored, compared to an event steam observing the same scene. 
Since our method operates on events which can capture very fast  motions 
without motion blur and data storage overheads, it can be used in  applications that involve tracking fast moving hands.
\begin{figure}[t!] 
\vspace{10pt}
\centering{ \includegraphics[width =0.5\textwidth]{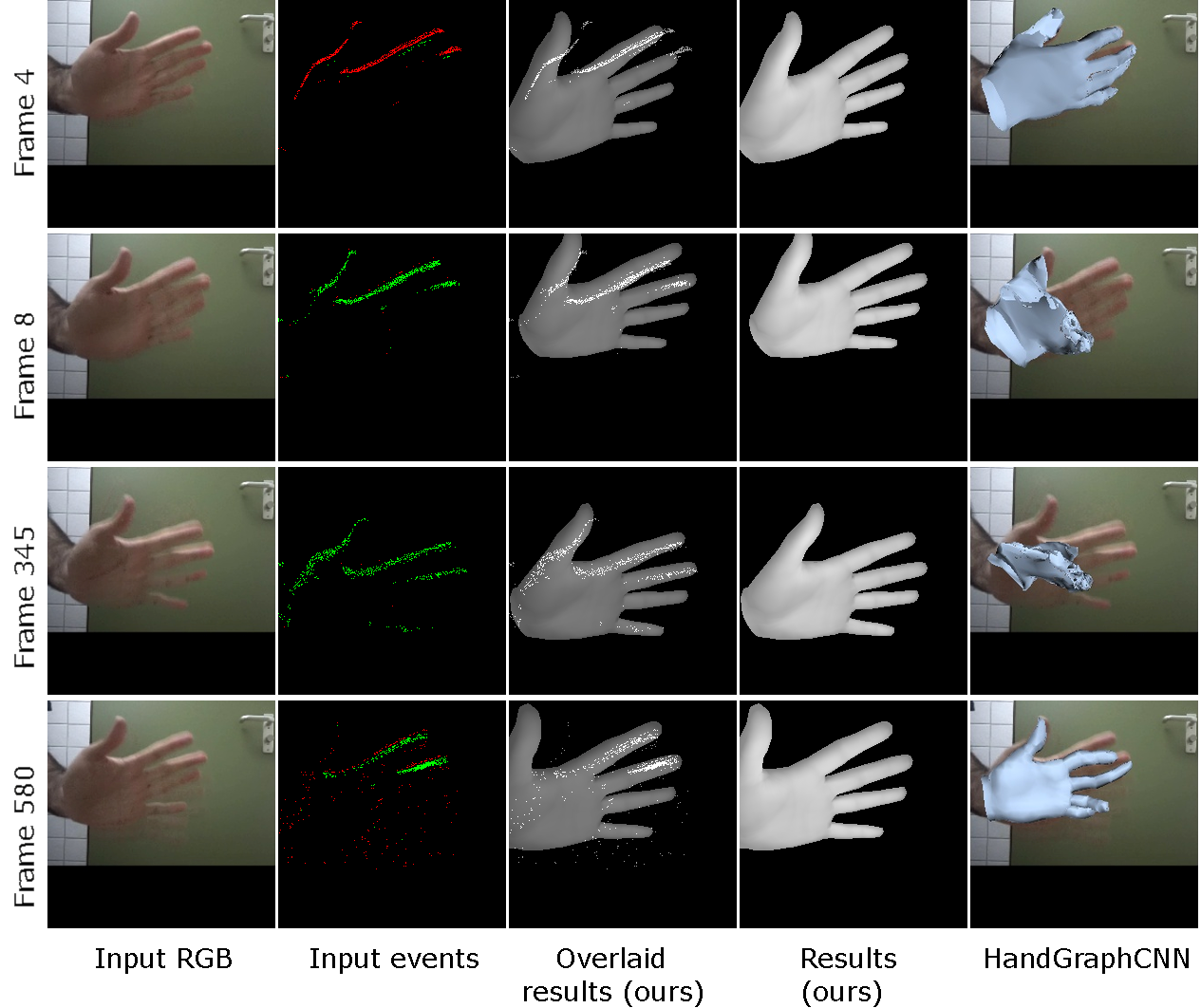}}
    \caption[Qualitative Results]{Results on the real fast hand sequence. Our method handles fast motion successfully.} 
    \label{fig:qualitativerealfasthand} 
\end{figure}
We record both RGB images and events of a fast-moving hand using Sony RX0 and DAVIS 240C, respectively. 
The cameras are time-synchronised using a flash and 
calibrated (intrinsically and extrinsically). 
Fig.~\ref{fig:qualitativerealfasthand} compares the qualitative results of our method against HandGraphCNN \cite{ge2019handshapepose} on fast hand waving motions, see our video for dynamic  visualisations. 
 Most RGB images suffer from motion blur and  HandGraphCNN \cite{ge2019handshapepose}---which is an RGB  image-based technique---often fails on the blurred images, as one would expect. 
In contrast, our method produces visually more plausible 3D reconstructions because it operates on events. 

\end{document}